\documentclass[10pt,twocolumn,letterpaper]{article}

\usepackage{cvpr}              
\newcommand{\colorRed}{\textcolor{red}}
\usepackage{multirow}

\usepackage{cuted}

\usepackage{colortbl}

\definecolor{baselinecolor}{gray}{.9}
\newcommand{\baseline}[1]{\cellcolor{baselinecolor}{#1}}

\usepackage{amsmath,amsfonts,bm}

\def\eqref#1{equation~\ref{#1}}

\def\1{\bm{1}}

\def\rvc{{\mathbf{c}}}
\def\rvd{{\mathbf{d}}}

\def\rvp{{\mathbf{p}}}

\def\rvt{{\mathbf{t}}}

\def\rvw{{\mathbf{w}}}

\def\rmG{{\mathbf{G}}}

\def\rmM{{\mathbf{M}}}

\DeclareMathAlphabet{\mathsfit}{\encodingdefault}{\sfdefault}{m}{sl}
\SetMathAlphabet{\mathsfit}{bold}{\encodingdefault}{\sfdefault}{bx}{n}

\definecolor{cvprblue}{rgb}{0.21,0.49,0.74}
\usepackage[pagebackref,breaklinks,colorlinks,allcolors=cvprblue]{hyperref}

\def\paperID{11466} 
\def\confName{CVPR}
\def\confYear{2025}

\title{RoMeO: \colorRed{Ro}bust \colorRed{Me}tric Visual \colorRed{O}dometry}

\author{
Junda Cheng$^{1,2}$\footnotemark[1],
~~Zhipeng Cai$^{2}$\footnotemark[1],
~~Zhaoxing Zhang$^{1}$,
~~Wei Yin$^{3}$,\\
~~Matthias Müller$^{2}$,
~~Michael Paulitsch$^{2}$,
~~Xin Yang$^{1}$\footnotemark[2]\\
[2mm]
$^1$~Huazhong University of Science and Technology \quad $^2$~Intel Labs \quad $^3$~Horizon Robotics\\
\tt\small \{Junda Cheng, zzx, xinyang2014\}@hust.edu.cn, czptc2h@gmail.com, \\ \tt\small matthias.mueller.2@kaust.edu.sa, yvanwy@outlook.com}

\begin{document}
\maketitle
\begin{abstract}
\renewcommand{\thefootnote}{\fnsymbol{footnote}}
\footnotetext[1]{Equal contribution. Work done during the internship at Intel Labs.}
\footnotetext[2]{Corresponding author.}
Visual odometry (VO) aims to estimate camera poses from visual inputs --- a fundamental building block for many applications such as VR/AR and robotics. This work focuses on monocular RGB VO where the input is a monocular RGB video without IMU or 3D sensors. Existing approaches lack robustness under this challenging scenario and fail to generalize to unseen data (especially outdoors); they also cannot recover metric-scale poses. 
We propose Robust Metric Visual Odometry (RoMeO), a novel method that resolves these issues leveraging priors from pre-trained depth models. RoMeO incorporates both monocular metric depth and multi-view stereo (MVS) models to recover metric-scale, simplify correspondence search, provide better initialization and regularize optimization. Effective strategies are proposed to inject noise during training and adaptively filter noisy depth priors, which ensure the robustness of RoMeO on in-the-wild data.
As shown in Fig.~\ref{fig:teaser}, RoMeO advances the state-of-the-art (SOTA) by a large margin \emph{across 6 diverse datasets covering both indoor and outdoor scenes}. Compared to the current SOTA DPVO, RoMeO reduces the relative (align the trajectory scale with GT) and absolute trajectory errors both by $>50\%$. The performance gain also transfers to the full SLAM pipeline (with global BA $\&$ loop closure). Code will be released upon acceptance at: \textcolor{magenta}{https://github.com/Junda24/RoMeO}.
\end{abstract}    
\section{Introduction}
\label{sec:intro}

\begin{figure}[t]
\centering
{\includegraphics[width=1.0\linewidth]{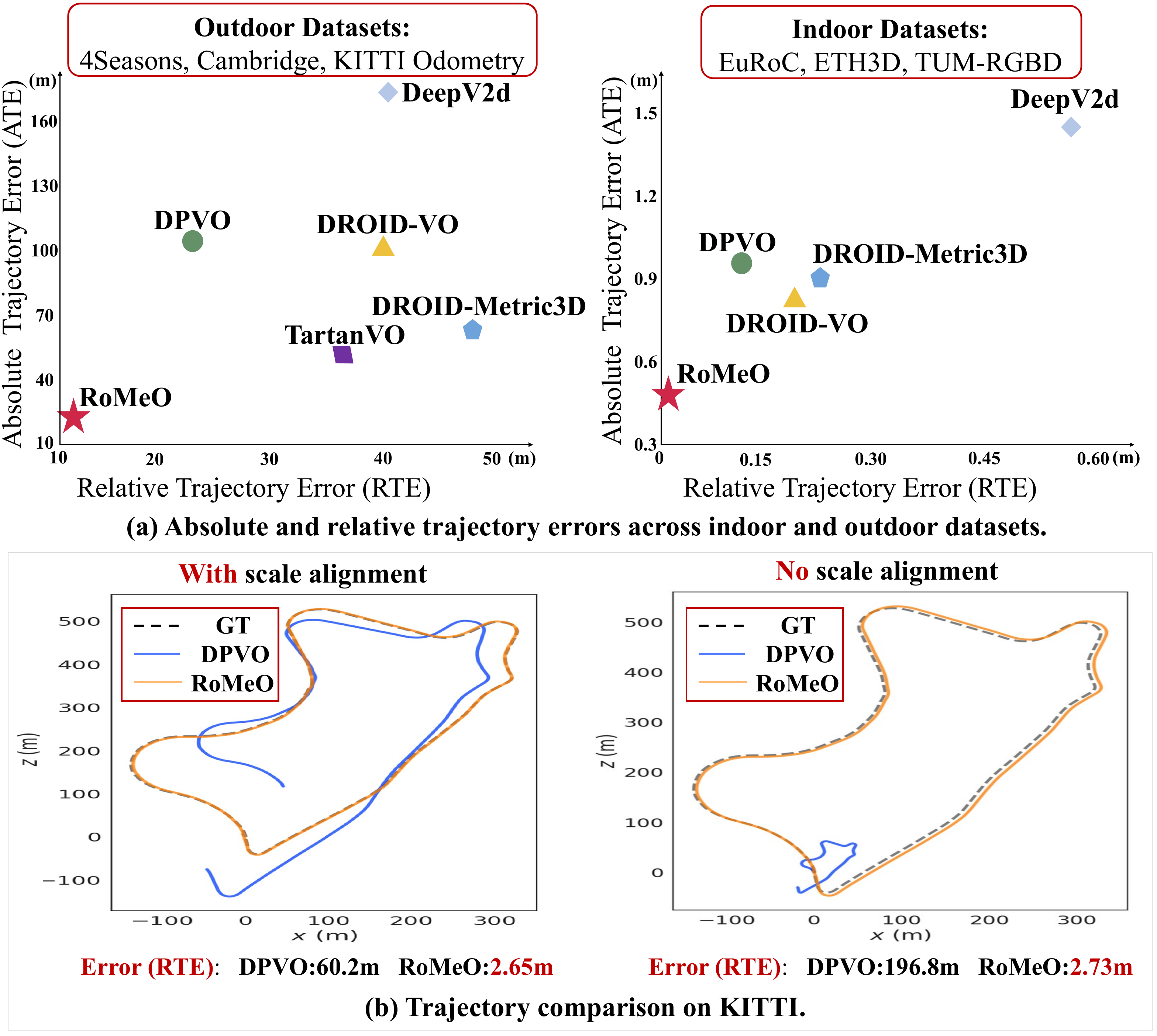}}
\vspace{-5pt}
\caption{\textbf{Teaser}.(a) RoMeO vs SOTA methods across 6 diverse datasets covering 3 indoor and 3 outdoor scenes, RoMeO outperforms SOTA methods by a large margin both in terms of the trajectory shape (relative trajectory error (RTE)) and scale (absolute trajectory error (ATE)). (b) RoMeO trajectories closely align with the ground truth, even without scale alignment.}
\label{fig:teaser}
\vspace{-15pt}
\end{figure}


Visual Odometry (VO) estimates sensor poses from visual signals. It is the core problem of many applications such as mapping, robot navigation, and autonomous driving. This work focuses on \emph{monocular RGB VO}, where the input is only a monocular RGB video and no information from a 3D sensor or inertial measurement unit (IMU) is available. 

Classical methods rely on hand-crafted features and explicit geometric optimizations to estimate poses. The optimization is executed either directly on pixel intensities~\citep{engel2014lsd, engel2017direct, zubizarreta2020direct}, or indirectly on matches between keypoints~\citep{mur2015orb, rosinol2020kimera}. 
Due to the instability of hand-crafted features, classical methods suffer from frequent tracking failures on challenging data with large motions or extreme weather.

Learning-based approaches~\citep{wang2021tartanvo, teed2021droid, zhang2024leveraging, teed2024deep} train end-to-end systems for both correspondence search and pose optimization, which minimizes tracking failures. However, these approaches have limited robustness, i.e., they generalize poorly on zero-shot data, where classical methods often perform better when they do not fail. Moreover, both types of methods lack mechanisms to recover metric-scale trajectories without 3D sensors or IMU. 


A common reason for these drawbacks is the lack of priors. As a result, the joint pose-depth optimization suffers from local optima and cannot recover the metric-scale. Though previous works try to use predicted depth~\citep{yang2020d3vo, yin2023metric3d} to improve VO, naive use of potentially noisy depth makes them generalize worse than pure correspondence-based SOTA~\citep{teed2021droid, teed2024deep}. To this end, we propose \emph{Robust Metric Visual Odometry (RoMeO)}, a novel method that can leverage potentially noisy priors from pre-trained depth models to improve VO accuracy, recover metric-scale poses and maintain robustness. 

RoMeO initializes VO with monocular metric depth models to recover metric scale poses. Robust depth-guided bundle adjustment adaptively detects accurate monocular and MVS priors, and uses them to regularize pose-depth optimization. Effective conditions are introduced to avoid poor MVS priors generated by inaccurate poses, low multi-view overlaps, or insufficient motions. Noise augmented training adapts the flow estimation network to the depth-enhanced inputs, which maximizes the accuracy while maintaining the robustness to depth prediction noise.


As shown in Fig.~\ref{fig:teaser} (a), RoMeO significantly advances the SOTA across 6 diverse benchmarks covering both indoor and outdoor scenes. Compared to the current SOTA DPVO~\citep{teed2024deep}, RoMeO achieves $>50\%$ reduction for both relative and absolute trajectory errors, i.e., both the trajectory shape and scale are significantly improved (Fig.~\ref{fig:teaser} (b)). Unlike previous depth-based approaches~\citep{yin2023metric3d} that hurt the VO accuracy on challenging data, the performance improvement of RoMeO is consistent across the board. 
These improvements also propagate to the full SLAM system with global BA enabled. 

\noindent\textbf{Contributions:}
(1) We devise a novel system that can effectively use both monocular and MVS depth priors to improve VO. (2) We introduce robust depth guidance into flow estimation and bundle adjustments, which significantly improves the accuracy even under noisy depth priors. (3) We conduct comprehensive experiments, validating the effectiveness of our method across diverse indoor and outdoor datasets, improving over SOTA by a large margin.

\section{Related work}\label{sec:related}

\noindent\textbf{Visual odometry.} Practical VO systems may rely on inputs beyond a monocular video, for example, IMU in visual-inertial odometry~\citep{forster2015imu}, sensor depth in RGB-D VO~\citep{whelan2013robust, teed2021droid, handa2014benchmark}, and multi-view sensors in stereo VO~\citep{wang2017stereo, engel2014lsd}. The focus of this work is VO from only monocular RGB video.

Classical methods use hand-crafted features to find correspondences and perform joint pose-depth optimization. Indirect approaches~\citep{mur2015orb, rosinol2020kimera} find correspondences through feature matching and then minimize the projection error on the correspondences. Direct approaches~\citep{engel2014lsd, engel2017direct, zubizarreta2020direct} minimize the photometric error directly without feature matching. A common drawback of classical methods is the frequent tracking failure on challenging data with large motions or extreme weather.

Learning-based methods train differentiable systems end-to-end on labeled data to avoid tracking failures. \cite{teed2018deepv2d} proposed a differentiable structure-from-motion architecture that alternates between motion and depth estimation. \cite{wang2021tartanvo} incorporated camera intrinsics into VO and pioneered training on the large-scale synthetic dataset TartanAir~\citep{wang2020tartanair}. \cite{teed2021droid} significantly improved prior works by introducing RAFT~\citep{teed2020raft} into VO systems and designing a differentiable bundle adjustment layer for pose-depth joint optimization. \cite{teed2024deep} sped up Droid-SLAM by sparse flow estimation and optimization. The major drawback of learning-based approaches is the limited robustness to zero-shot data, especially for outdoor scenes with challenging motion and dynamic objects. As a result, they can perform worse than classical methods which fail less in such cases. Meanwhile, existing approaches cannot recover metric-scale poses without 3D sensors (depth, stereo, or IMU). Though several works~\citep{tateno2017cnn, yang2020d3vo, yin2023metric3d} tried to introduce predicted depth to VO, they suffer from performance drop due to the depth prediction noise, especially on zero-shot data. RoMeO improves VO by introducing depth priors from pre-trained monocular and MVS models to both the initialization and the iterative optimization of VO. Effective strategies are proposed to ensure the robustness of RoMeO under (severe) prior noise, which enables consistent and significant error reduction across diverse zero-shot data. 

\noindent\textbf{Depth estimation.} Deep learning-based depth estimation methods~\citep{bhat2023zoedepth, yin2023metric3d, yang2024depth, feng2024mc, cheng2024adaptive, zhu2025svdc, cheng2024coatrsnet, cheng2022region, cheng2025monster, ranftl2020towards,ranftl2021vision} have demonstrated superior performance on commonly used benchmarks. \emph{Monocular depth} estimation aims to recover depth from a single image. Recent methods~\citep{ranftl2020towards,ranftl2021vision, bhat2023zoedepth, yin2023metric3d, yang2024depth} have trained models on large-scale data to enable zero-shot monocular depth estimation. \emph{Multi-view stereo (MVS)} methods~\citep{yao2018mvsnet, cheng2024adaptive, bae2022multi, cao2022mvsformer} estimate depth from posed multi-view images. With accurate poses for the input images, they can recover more consistent and accurate depth than monocular methods. RoMeO leverages both monocular and MVS models to better initialize and regularize VO.

\section{Method}
\label{sec:Method}

\begin{figure*}[t]
  \centering
  \includegraphics[width=0.98\textwidth]{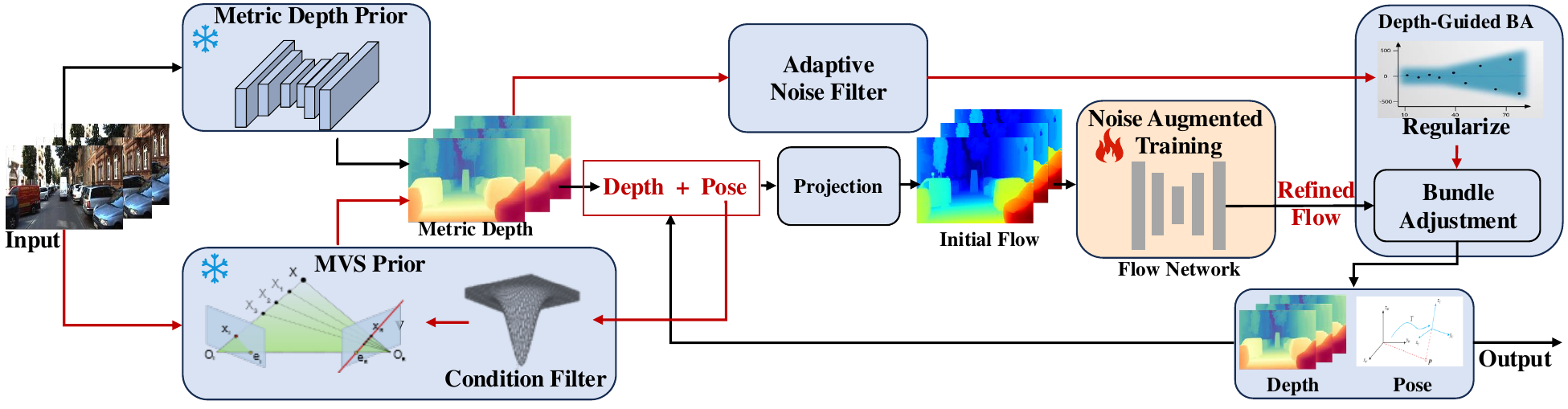}
  \caption{\textbf{Overview}. RoMeO initializes each frame using monocular metric depth models. MVS models are used to further refine intermediate BA depth. Besides replacing the initial/intermediate depth, monocular and MVS depth priors are also added into the regularization terms of BA, with adaptive conditions to filter noisy depth priors and enable effective MVS prediction. Noise augmented training is used to adapt the flow network to depth-enhanced inputs, which maximizes the accuracy while maintaining the robustness to prior noise.
  }
  \label{fig:network}
\end{figure*}

As shown in Fig.~\ref{fig:network}, RoMeO maintains a frame graph during VO, which contains a sliding window of keyframes. Every new frame is initialized with an identity relative pose and the predicted monocular \emph{metric} depth from a pre-trained model. 
Given the current frame graph, RoMeO jointly refines the poses and depth by alternating between \emph{flow estimation} and \emph{differentiable bundle adjustment (BA)}, with robust guidance from both monocular and MVS depth.


Inspired by the recent SOTA~\citep{teed2021droid, teed2024deep}, each \emph{flow estimation} step first obtains an initial optical flow by projecting the pixels of each keyframe to the others using camera intrinsics and the current pose and depth. This initial flow is then fed to a flow network that predicts a \emph{residual flow} and a confidence map. RoMeO leverages depth priors to obtain more accurate initial flow, which simplifies residual flow prediction. Any flow model can in principle be used in this step. Due to the wide adoption and the ability to perform dense reconstructions, we use a RAFT-style architecture following~\citep{teed2021droid}. After flow estimation, 2 iterations of \emph{BA} are performed to optimize the pose and depth. RoMeO leverages monocular metric depth to initialize BA such that it can recover metric scale poses without 3D sensor or IMU. Both monocular and MVS depth are further used to effectively regularize the BA objective. This process is repeated 6 times, resulting in 6 flow estimation steps and 12 BA iterations for each new keyframe. Pose-only BA is performed for non-keyframes, with monocular depth initialization.

\subsection{Robust Depth-Guided Bundle Adjustment}\label{sec:method_MVS}




RoMeO guides BA with both monocular and MVS depth. Besides using them to replace initial/intermediate BA depth, it embeds them into the BA objective below:
\begin{equation}\label{eq:penalty}
    \begin{split}
        \underset{\rmG, \rvd}{\text{min}} & \underset{(i,j) \in \varepsilon}{\sum} (\|\rvp_{ij}^* - \Pi_c(\rmG_{ij}\circ \Pi_c^{-1}(\rvp_i, \rvd_i)) \|^2_{\Sigma_{ij}} + \\
        & C_i \lambda \|\rvd_i - \rvd^*_i\|^2_{\Sigma_{ij}}),
    \end{split}
\end{equation}
where $\rmG$ and $\rvd$ are the optimized poses and depth, and $\epsilon$ is the current frame graph. The first term is the standard BA objective~\citep{teed2021droid} where $\rvp_{ij}^* \in \mathbb{R}^{H \times W \times 2}$ denotes the image coordinates when we project the pixels from frame $i$ to $j$ \emph{with the estimated flow}. $\Pi_c(\rmG_{ij}\circ \Pi^{-1}(\rvp_i, \rvd_i))$ corresponds to the image coordinates when we project the pixels from frame $i$ to $j$ \emph{with the current depth $\rvd_i$ of frame $i$, the relative pose $\rmG_{ij}$ between frames $i$ and $j$}. $\Pi_c$ is the world-to-camera projection. Intuitively, the first term encourages the consistency between flow-based and geometric-projection-based correspondences. The second term is a robust depth regularization term where we encourage the BA-optimized depth map $\rvd_i \in \mathbb{R}^{H \times W}$ to be close to the predicted monocular/MVS depth $\rvd_i^*$. $\lambda = 0.05$ is the regularization weight. The condition weight $C_i \in \{0, 1\}$ turns off depth regularization when severe noise exists in $\rvd_i^*$. $\|\cdot\|_{\Sigma_{ij}}$ is the Mahalanobis distance which weights the error terms based on the confidence $\rvw_{ij} \in [0,1]^{H \times W}$ of flow estimation.

RoMeO automatically determines the value of $C_i$, which is the key to making depth-guided BA robust to prior noise. During the construction of the initial frame graph containing the first 12 input keyframes, we do not enable depth regularization ($C_i = 0$). After the initial frame graph has been built and the pose-depth optimization is completed on this graph, we compute the average photometric error from the latest keyframe $i$ to the connected frames in the graph:
{\small
\begin{equation}\label{eq:adaptive_threshold}
        \eta(i) = \underset{(i,j) \in \varepsilon'}{\sum} ||\rvc_{i}(\rvp_i) - \rvc_j(\Pi_c(\rmG_{ij}\circ \Pi_c^{-1}(\rvp_i, \rvd_i))) ||^2_{\Sigma_{ij}},
\end{equation}
}
where $\rvc_i(\rvp_i) \in \mathbb{R}^{H \times W \times 3}$ returns the colors of the pixels in frame $i$. Eq.~\ref{eq:adaptive_threshold} essentially computes the color difference between the pixels in frame $i$ with their corresponding pixels in frame $j$ where the correspondences are obtained by the same pose-depth re-projection in Eq.~\ref{eq:penalty}. When the depth of frame $i$ is accurate, the photometric error $\eta(i)$ should be small. We denote the error on the latest keyframe as $\eta_\text{init}$. 

For every new keyframe in the future, we compute the same photometric error $\eta'$ using Eq.~\ref{eq:adaptive_threshold}, where the depth is the initial monocular depth, and the pose is obtained from the first BA iteration. We only set $C_i$ to 1 if $\eta' < \alpha \eta_{\text{init}}$, where $\alpha$ is a pre-defined constant. Intuitively, when this condition is violated, the predicted depth is potentially unreliable and should not be used to regularize BA. This strategy is robust to noise: it not only maintains the strong performance gain from accurate depth priors but also mitigates the negative impacts of severe depth noise. As a result, RoMeO achieves much higher zero-shot accuracy than current SOTA across a diverse set of datasets.


\subsection{Metric Depth Prior}

The non-convex nature makes VO optimization highly sensitive to initialization. State-of-the-art methods often initialize the depth of a new frame to a constant (e.g. 1m)~\citep{teed2021droid} for all pixels, which might be acceptable for indoor scenes where the depth variation is small. However, it is unsuitable for outdoor scenes where close and distant pixels can have highly different depths. Uniform initialization also hinders the recovery of metric-scale poses. To address both problems, we initialize the depth of each frame with the output of a pre-trained monocular \emph{metric} depth model. 

\begin{table}[!htb]
  \centering
  \caption{\textbf{Different monocular depth models for initialization and depth guided BA.} Other strategies mentioned in later sections are not applied.}
  \label{tab:depth_compare}
  \resizebox{1\columnwidth}{!}{
  \begin{tabular}{@{}l|c|c|c|c@{}}
    \toprule
    \multirow{2}{*}{Model} & \multicolumn{2}{c|}{KITTI Odometry}  & \multicolumn{2}{c}{TUM-RGBD} \\
    \cmidrule{2-3} \cmidrule{4-5}
    & RTE (m) / ATE (m) & FPS & RTE (m) / ATE (m) & FPS \\
    \midrule
    No depth prior  & 47.53/137.33  & \textbf{5.33} & 0.116/0.551 & \textbf{10.77} \\
    \midrule
    DepthAnythingV2-Small & 11.15/18.78 & \underline{4.19} & \underline{0.104}/0.273 & 7.56\\
    DepthAnythingV2-Large & 7.97/13.47 & 1.59  & 0.107/0.456 & 4.20\\
    Metric3DV2-Small & 5.71/8.74 & 4.08  & \underline{0.104/0.235} & 6.41\\
    Metric3DV2-Large & \textbf{4.08/5.73} & 1.70  & 0.105/0.247 & 2.39\\
    DPT-Hybrid & \underline{4.25/8.54} & 3.96  & \textbf{0.098/0.205} & \underline{8.55}\\
    \bottomrule
  \end{tabular}
  }
\end{table}
Though more recent methods~\citep{hu2024metric3d, yang2024depth} with large pre-trained models are available, they introduce severe overheads to the VO system and hinder the practicality. To minimize overhead while maintaining robustness, we use the lightweight DPT-Hybrid~\citep{ranftl2021vision} with the provided scale and shift parameters to obtain the metric depth. Tab.~\ref{tab:depth_compare} shows the accuracy and speed of VO using different metric depth models for initialization and BA regularization. We use 1 indoor (KITTI) and 1 outdoor (TUM-RGBD) datasets for analysis. Interestingly, though all depth models can provide reasonable error reduction, bigger models such as DepthAnythingV2-Large~\citep{yang2024depth} and Metric3DV2-Large~\citep{hu2024metric3d} introduce severe overheads to VO, and are not obviously better than DPT-Hybrid. We conjecture that this is because the BA of RoMeO will optimize the depth details and filter out noisy depth priors. Therefore, the key to practical monocular depth guidance in VO is a light-weight model that can provide rough but robust metric-scale depth.

\subsection{MVS Prior}


Creating effective MVS guidance in a VO system is non-trivial for three main reasons: (1) \emph{Accurate MVS requires a reasonable amount of camera translations and rotations, and large overlapping areas between multiple views.} These conditions are not ensured by standard keyframe selection strategies. (2) \emph{Accurate MVS needs accurate poses.} Naively adding MVS guidance at early iterations can hurt VO. (3) \emph{MVS outputs often contain noise.}

To address the first issue, we denote the intermediate relative translation estimated by BA between the last 3 keyframes ($i-2, i-1, i$) as ${\rvt_{i-2,i-1}}$ and ${{\rvt}_{i-1, i}}$, and only enable MVS guidance when

\begin{equation}\label{eq:trans_cond}
\begin{aligned}
&\|{{\rvt}_{k-1,k}}\| + \|{{\rvt}_{k-2,k-1}}\| > 0.1m \And\\
&\angle({{\rvt}_{k-2,k-1}}, {{\rvt}_{k-1,k}}) \in [10^\circ, 30^\circ].
\end{aligned}
\end{equation}

This strategy can effectively filter keyframes with extreme motions or small cross-view overlaps. 

\begin{figure}[!htb]
  \vspace{-1em}
  \centering
  \includegraphics[width=.8\columnwidth]{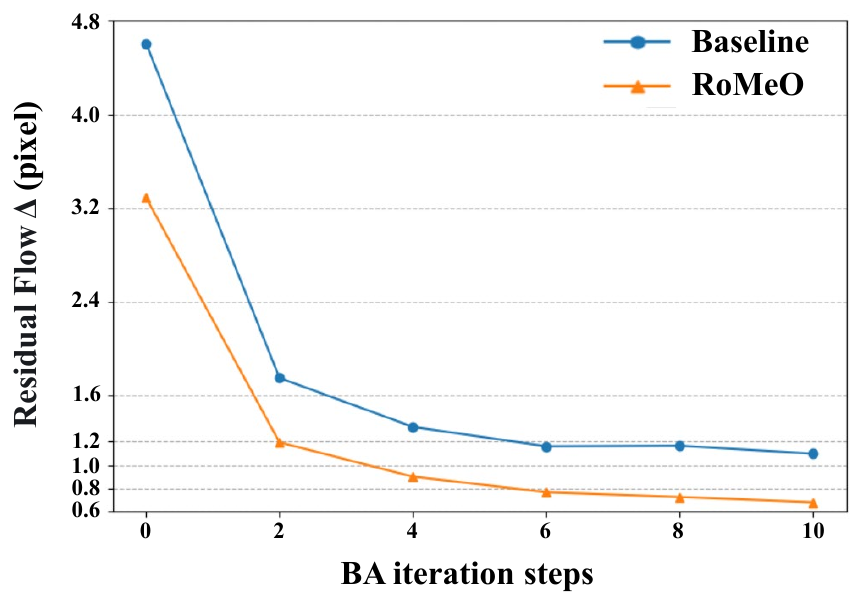}
  \vspace{-10pt}
  \caption{\textbf{Residual flow magnitudes in different BA iterations.}}
  \label{fig:flow_res_2}
  \vspace{-1em}
\end{figure}

For the second issue, an interesting observation is that, given effective monocular depth guidance, BA can return reasonably accurate intermediate poses within a small number of iterations. This is evident from Fig.~\ref{fig:flow_res_2} where we plot the magnitude of the residual flow over different BA iterations for the \emph{KITTI Odometry} dataset~\citep{Geiger2012CVPR}. The magnitude of RoMeO at early iterations is already lower than the final iteration of the baseline without any depth prior, which indicates better pose accuracy. Based on this observation, if E.q.~\ref{eq:trans_cond} is satisfied after the 8-th BA iteration, RoMeO uses the intermediate poses and the most recent 3 keyframes to compute the MVS depth for the newest keyframe. We use pre-trained MaGNet\citep{bae2022multi} as the MVS estimator. The intermediate BA depth and $\rvd_i^*$ in E.q.~\ref{eq:penalty} are then replaced with the MVS depth to improve the remaining flow generation and pose-depth optimization accuracy.

To address the third issue, we leverage the output confidence map $\rmM \in \mathbb{R}^{H \times W}$ of MVS, which indicates the probability that the prediction on each pixel is correct. Specifically, we ignore the pixels corresponding to the lowest $20\%$ of the values in $\rmM$ during future flow generation and BA iterations. Essentially, sparse BA is conducted on selected pixels with reliable depth.

\begin{figure*}[ht]
  \centering
  \includegraphics[width=0.87\textwidth]{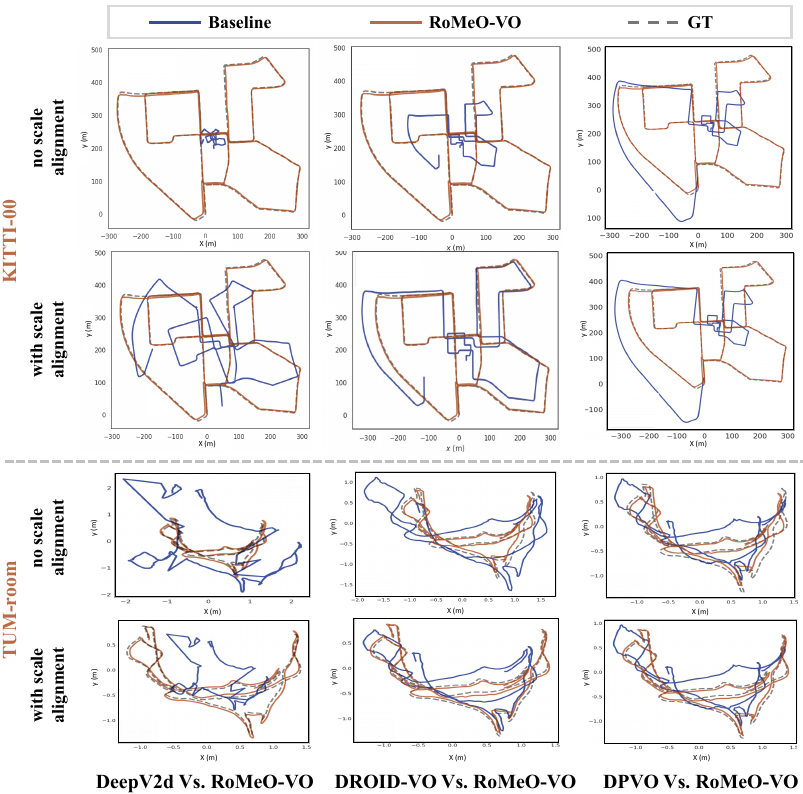}
  \vspace{-0.5em}
  \caption{\textbf{Trajectory}. For each scene, the two rows show respectively the trajectory with and without scale alignment. RoMeO aligns much better with GT both with and without scale alignment.}
  \label{fig:traj}
  \vspace{-1.2em}
\end{figure*}

\subsection{Noise Augmented Training}\label{sec:Joint_training}

The trainable parameters of RoMeO lie in the flow network. Besides benefiting BA optimization, accurate depth priors effectively simplifies flow estimation since the initial flow obtained by pose-depth based reprojection would be more accurate. Empirically we can train the flow network without depth priors, and insert depth priors during inference to achieve non-trivial performance improvements. However, this naive strategy cannot fully adapt the trainable parameters to the prior-enhanced inputs. Specifically, our prior-enhanced poses and depth lead to more accurate initial flow, which requires a much smaller residual flow during inference. Hence, we jointly train the flow network with the depth-enhanced inputs. 

Since predicted depth can have (severe) noise on unseen data, we purposely maintain some noise during training to ensure the robustness to prior noise. Specifically, we first pre-train the flow network without any depth prior. Then, we fine-tune the pre-trained system with monocular depth initialization with a much smaller learning rate. MVS depth is not used during fine-tuning to ensure a reasonable amount of depth noise. 

The fine-tuning is done on \emph{TartanAir}~\citep{wang2020tartanair}, the same simulation dataset used for pre-training. There is a strong domain gap between simulation and real data, for example, some TartanAir scenes contain extreme depth values rarely seen in real-world data (e.g., objects that are $16km$ away). The monocular depth models often return heavily erroneous predictions in these cases, making fine-tuning ineffective. To address this problem, we align the scale and shift of the predicted depth to the ground truth when the relative prediction error exceeds $20\%$. This strategy minimizes the negative impact of domain shift, making the flow network adapt to more accurate depth in practice. Meanwhile, only aligning predictions with large errors ensures that the fine-tuning data still contain a reasonable amount of noise.

With effective noise handling strategies, the depth prior in RoMeO significantly enhances flow generation since the cross-view re-projection is much more accurate. As supported by Fig.~\ref{fig:flow_res_2}, a strong depth prior leads to much smaller flow adjustment during iterative refinement, i.e., reliable correspondences are obtained much faster. A strong depth prior also helps BA to avoid bad local optima, making it easier to find accurate poses. These advantages create a \emph{positive feedback loop}, where better depth leads to better flow and poses, which further enhance depth optimization.

\begin{table*}[ht]
  \caption{\textbf{Visual Odometry Evaluation}. Results are reported as RTE (m) (with scale alignment) / ATE (m). Missing values (--) indicate that the method loses track in some sequences.  ORB-SLAM3 (VO) means the VO front-end of ORB-SLAM3. RoMeO outperforms both learning-free and learning-based baselines across the board. Both the metric-scale (ATE) and trajectory shape (RTE) improve significantly.
  }\label{tab:result_VO_main}
  \centering
  \resizebox{1\linewidth}{!}{
  \begin{tabular}{@{}l|cccc|cccc@{}}
  \toprule
     \multirow{2}{*}{Dataset} & \multicolumn{4}{c|}{Outdoor} & \multicolumn{4}{c}{Indoor} \\
     \cline{2-9}
     & KITTI Odometry & 4Seasons & Cambridge & Avg error & EuRoC & ETH3D & TUM-RGBD & Avg error\\
    \hline
    ORB-SLAM3 (VO)  & 29.10/163.25  &-- & -- & -- & 0.488/1.677 & --  & -- & -- \\
    DSO  & 47.23/154.25 & -- & -- & -- & 0.404/1.532 & --   & -- & --\\
    DeepV2d  & 22.20/154.52 &74.70/344.75 & 27.21/34.50 & 41.37/177.92 & 1.173/3.452 & 0.324/0.582 & 0.225/0.415 & 0.574/1.483 \\
    TrianFlow  & 42.07/168.17 & 40.44/150.46 & 27.35/34.60 & 36.62/117.74 & 1.731/1.848  & 0.706/0.868 & 0.444/0.565 & 0.960/1.094\\
    TartanVO  & 32.25/45.30 & 59.25/\underline{81.70} & 19.40/\underline{24.23} & 36.97/\underline{50.41} & 0.632/4.135 & 0.421/26.273 & 0.320/16.692 & 0.458/15.70 \\
    DROID-VO & 47.53/137.33 & 58.87/149.54 & 13.56/33.67 & 39.99/106.85 & 0.141/\underline{1.307} & 0.367/\underline{0.628} & 0.116/0.551 & 0.208/\underline{0.829}\\   
    DROID-Metric3d & \underline{3.95/5.57} & 125.44/140.32 & \underline{13.55}/47.68 & 47.64/64.52 & 0.109/1.310 & 0.420/1.077 & 0.190/0.330 & 0.240/0.906 \\
    DPVO & 46.04/140.28 & \textbf{9.95}/141.36 & 15.89/36.03 & \underline{23.96}/105.89 &\underline{0.101}/1.865 & \underline{0.203}/0.646 & \underline{0.107/0.324} & \underline{0.137}/0.945 \\
    \baseline{RoMeO-VO(Ours)} & \baseline{\textbf{2.71/3.81}} & \baseline{\underline{19.59}/\textbf{42.56}} & \baseline{\textbf{9.96/23.47}} & \baseline{\textbf{10.75/23.28}}  &  \baseline{\textbf{0.098/1.126}} & \baseline{\textbf{0.022/0.238}} & \baseline{\textbf{0.067/0.091}} & \baseline{\textbf{0.062/0.485}} \\
  \bottomrule
  \end{tabular}
  }
\end{table*}

\begin{table*}[ht]
  \caption{\textbf{SLAM Evaluation}. Results are reported as RTE (m) (with scale alignment) / ATE (m). Missing values (--) indicate that the method loses track in some sequences. The performance gain of RoMeO on VO propagates to the full SLAM system.
  }
  \centering
  \resizebox{1\linewidth}{!}{
  \label{tab:result_SLAM_main}

  \begin{tabular}{@{}l|cccc|cccc@{}}
    \toprule
     \multirow{2}{*}{Dataset} & \multicolumn{4}{c|}{Outdoor Dataset} & \multicolumn{4}{c}{Indoor Dataset} \\
     \cline{2-9}
     & KITTI Odometry & 4Seasons & Cambridge & Avg error & EuRoC & ETH3D & TUM-RGBD & Avg error\\
    \hline
    ORB-SLAM3   & 16.42/165.28  & -- & -- & -- &  0.214/1.437 & -- & -- & --\\
    Droid-SLAM & 39.12/155.89 & 58.12/144.20 &12.37/33.13 & 36.54/111.07 & 0.019/1.238 & 0.010/\textbf{0.011} & 0.028/0.168 & 0.019/0.472 \\
    \baseline{RoMeO-SLAM} & \baseline{\textbf{2.64/3.72}} & \baseline{\textbf{20.39/44.20}}  & \baseline{\textbf{9.90/23.23}} & \baseline{\textbf{10.98/23.72}} & \baseline{\textbf{0.016/1.100}} & \baseline{\textbf{0.008}/0.060} & \baseline{\textbf{0.021/0.071}} & \baseline{\textbf{0.015/0.410}}\\
  \bottomrule
  \end{tabular}
  }
\end{table*}

\section{Experiments}\label{sec:exp}

\noindent\textbf{Implementation.} 
RoMeO is implemented using a mixture of PyTorch and C++ following the official code of~\citep{teed2021droid}. To maximize the performance, we use separate depth models/hyperparameters for indoor and outdoor scenes. For depth initialization, we use DPT-Hybrid with the provided scale and shift hyperparameters (scale=0.000305, shift=0.1378 for indoor and scale=0.00006016, shift=0.00579 for outdoor)~\citep{DPT_hyperparam}. For MVS guidance, we use the corresponding MaGNet~\citep{bae2022multi} models for indoor and outdoor scenes. We set $\alpha$ in E.q.~\ref{eq:adaptive_threshold} to $1.75$ and $1.5$ respectively for outdoor and indoor scenes. Note that although the depth models/hyperparameters are different, we only train a \emph{single} VO system and use it for both indoor and outdoor scenes. During RoMeO fine-tuning, we only use the outdoor DPT scale and shift to predict the initial depth, since TartanAir is mostly outdoor. The improper scale and shift will be re-aligned with GT anyway when they create large prediction errors. The
initial learning rate of fine-tuning is reduced to 0.0001, with other setups the same as pre-training. Both stages of noise augmented training require roughly 7 days on 4 RTX-3090 GPUs. 

\noindent\textbf{Data.} Previous monocular RGB VO systems are mostly evaluated on indoor scenes. We use 6 zero-shot datasets with diverse indoor and outdoor scenes to thoroughly evaluate the robustness of different algorithms:
\begin{itemize}
    \item \emph{KITTI Odometry}~\citep{Geiger2012CVPR}: an outdoor self-driving dataset.
    \item \emph{4Seasons}~\citep{wenzel20214seasons}: an outdoor driving dataset with diverse scenes and weather conditions. 
    \item \emph{Cambridge}~\citep{kendall2015posenet}: a large-scale outdoor visual localization dataset taken around Cambridge University using a handheld camera. 
    \item \emph{EuRoC}~\citep{Burri25012016}: an indoor dataset captured by a Micro Aerial Vehicle (MAV).
    \item \emph{TUM-RGBD}~\citep{sturm2012benchmark}: an indoor dataset captured with a handheld camera. 
    \item \emph{ETH3D}~\citep{schops2019bad}: we use the SLAM benchmark of ETH3D, which is an indoor dataset with LiDAR captured depth. 
\end{itemize}
Since \emph{4Seasons} and \emph{ETH3D} contain too many sequences, we randomly select 1 training sequence of each scene for evaluation, see Appendix~\ref{appdx:detailed_results} for details.

\noindent\textbf{Metrics.} Following~\cite{campos2021orb}, we evaluate different methods using the \emph{Absolute Trajectory Error (ATE)}. To distinguish the metric scale accuracy and the trajectory shape accuracy, we define an additional metric where we align the scale of the output trajectory with the ground truth (GT); we call this metric \emph{Relative Trajectory Error (RTE)}. Note that previous monocular VO papers~\cite{teed2021droid, teed2024deep} report RTE as ATE since they cannot recover metric scale poses and by default assume scale alignment with GT.

\subsection{Main Results}\label{sec:exp_main}

\begin{table*}[ht]
\vspace{-0.5em}
  \caption{\textbf{Ablation study}. Removing or replacing RoMeO components hurts the performance (RTE/ATE). Noise augmented training is abbreviated as NAT.}
  \label{tab:ablation}
  \centering
  \resizebox{\linewidth}{!}{
  \begin{tabular}{@{}l|c|c|c|c|c|c|c@{}}
    \toprule
    Model & Mono depth & NAT & MVS & KITTI Odometry & 4Seasons & TUM-RGBD & Mean RTE/ATE \\
    \midrule
    RoMeO-VO (ours) & DPT & \checkmark & \checkmark &  2.71/3.81 & 19.59/42.56 & 0.067/0.091 & \textbf{7.45/15.49} \\
    always regularize depth & DPT & \checkmark & \checkmark & 2.48/3.63 & 117.95/154.78  & 0.136/0.158 & 40.20/52.86 \\
    no depth regularization & DPT & \checkmark & \checkmark & 11.25/47.91 & 16.99/39.63 & 0.058/0.088 & 9.44/29.21 \\
    DPT $\rightarrow$ Metric3D & Metric3D & \checkmark & \checkmark & 3.23/4.67 & 53.43/89.28 & 0.135/0.288 & 18.94/31.41 \\
    no MVS & DPT & \checkmark & &  3.35/5.14 & 23.49/55.38 & 0.075/0.121 & 8.96/20.21 \\
    no NAT \& no MVS & DPT & & & 4.25/8.54 & 34.32/121.65 & 0.098/0.205 & 12.89/43.47 \\
    no depth prior & & & & 47.53/137.33 & 58.87/149.54 & 0.116/0.551 & 35.52/95.79 \\
    \bottomrule
  \end{tabular}
  }
\vspace{-0.5em}
\end{table*}

We first compare RoMeO with state-of-the-art methods. Since RoMeO can be used as a standalone VO system and also as part of a full SLAM pipeline, we conduct experiments for both applications. 

Table~\ref{tab:result_VO_main} shows the comparison to VO systems. Learning-free baselines (ORB-SLAM3 (VO) and DSO) fail frequently on challenging data due to the instability of hand-crafted features. ORB-SLAM3 (VO) and DSO have $100\%$ success rates only on \emph{KITTI Odometry} and \emph{EuRoC}. Appendix~\ref{appdx:detailed_results} further reports results on individual sequences of each dataset. Learning-based baselines rarely lose track due to the improved stability of the end-to-end framework. However, the trajectory accuracy is limited especially on outdoor scenes. Most of them perform worse than ORB-SLAM3 (VO) on \emph{KITTI Odometry}. The closest baseline to RoMeO is DROID-Metric3d, where Metric3d~\citep{yin2023metric3d} depth is applied to the initialization and BA regularization of DROID-VO. Our implementation uses the original code from the Metric3d authors. The Metric3d paper showed significant performance improvement on \emph{KITTI Odometry}, which was consistent with our experiment. However, the same strategy can hurt the performance on other datasets (e.g., RTE on \emph{4Seasons} and \emph{TUM-RGBD}, ATE on \emph{Cambridge} and \emph{ETH3D}). 

RoMeO significantly and consistently outperforms both learning-based and learning-free baselines. Compared to the current SOTA DPVO, RoMeO reduces RTE and ATE respectively by $55.2\%$ and $77.8\%$ on average, and $>90\%$ on challenging data, e.g., the ATE of \emph{KITTI Odometry} improves from $140.28m$ to $3.81m$. This shows that RoMeO can significantly improve both the trajectory scale and shape, and generalizes to both indoor and outdoor scenes. 

\begin{figure*}[ht]
  \centering
  \includegraphics[width=1.\textwidth]{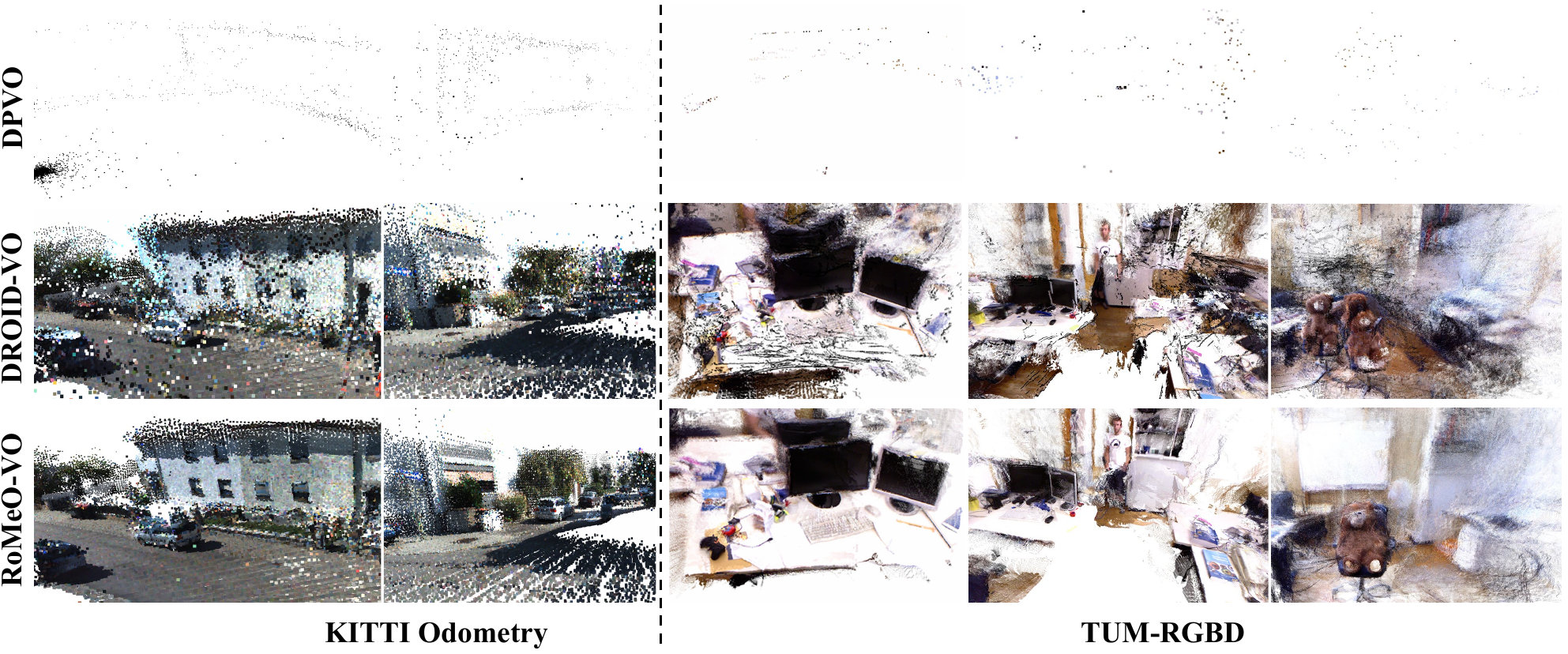}
  \vspace{-0.5em}
  \caption{\textbf{Point cloud visualization}. \emph{First two columns}: results on KITTI Odometry. \emph{Last three columns}: results on TUM-RGBD. RoMeO provides dense and more accurate 3D reconstructions.}
  \label{fig:point_cloud}
\vspace{-1.2em}
\end{figure*}

Fig.~\ref{fig:traj} shows the trajectory visualizations. Consistent with the quantitative results, the predicted trajectory of RoMeO aligns better with GT both with and without scale alignment. Moreover, it is the only method that can close the loop without applying global bundle adjustment or loop closure. Fig.~\ref{fig:point_cloud}  shows the visualization of reconstructed point clouds. For both indoor and outdoor scenes, RoMeO provides dense reconstructions of much higher quality. The reconstruction of DPVO is extremely sparse since it applies sparse flow estimation to accelerate VO.

Table~\ref{tab:result_SLAM_main} shows the comparison to SLAM systems. We build RoMeO-SLAM with global bundle adjustment enabled. Similar to the case of VO, RoMeO effectively improves the accuracy of the full SLAM system. The improvement level remains similar as in the VO case, e.g., $93.3\%$ and $97.6\%$ reduction rate on \emph{KITTI Odometry} for RTE and ATE respectively.

\subsection{Analysis}\label{sec:exp_analysis}


In this section, we provide a detailed method analysis. We first conduct ablation studies to validate each RoMeO component. Due to the computation cost, the ablation study is conducted on 3 evaluation datasets (1 indoor and 2 outdoor). The results are reported in Table~\ref{tab:ablation}.

\noindent\textbf{Depth guided BA.} To verify the effectiveness of our depth regularization in BA (E.q.~\ref{eq:penalty}), we report in row 2 (\emph{always regularize depth}) and 3 (\emph{no depth regularization}) of Table~\ref{tab:ablation} the performance of RoMeO with depth regularization always enabled (whether the predicted depth is accurate or not) and always disabled. When the depth regularization is always enabled, the performance of RoMeO drops severely on \emph{4Seasnos} ($19.59m \rightarrow 117.95m$ for RTE) due to the negative impact of noisy monocular/MVS depth. On the other hand, when the depth regularization is always disabled, the performance on \emph{KITTI Odometry} drops heavily ($3.81m \rightarrow 47.91m$ for ATE) due to the lack of effective BA regularization. With adaptive noise filtering, RoMeO maintains most performance gain from depth regularization, and minimizes the negative impact from inaccurate monocular/MVS depth, performing robustly across different datasets.


\noindent\textbf{MVS prior.} RoMeO applies MVS depth to enhance the intermediate flow and pose estimation. As shown in row 5 of Table~\ref{tab:ablation} (\emph{no MVS}), removing MVS priors hurts both ATE and RTE, showing the importance of the proposed module.

\noindent\textbf{Noise augmented training.} Noise augmented training plays a crucial role in adapting the VO network to the depth-prior-enhanced inputs. 
As shown in row 6 of Table~\ref{tab:ablation} (\emph{no NAT $\&$ no MVS}), removing noise augmented training from \emph{no MVS}, i.e., only using the pre-trained flow model in RoMeO, further worsens both ATE and RTE. E.g., in \emph{4Seasons}, the ATE increases by $>2$x. This result demonstrates the importance of fine-tuning to maximize the performance gain of RoMeO. 


\noindent\textbf{Metric depth prior.} Depth initialization enables not only better flow initialization but also the ability to obtain accurate metric-scale poses. The last row of Table~\ref{tab:ablation} (\emph{no depth prior}) removes depth initialization from \emph{no NAT $\&$ no MVR}, leading to a drastic error increase. On \emph{KITTI Odometry}, both ATE and RTE increase by more than $10$x, showing that the model cannot maintain a reasonable metric scale and accurate trajectory shape without depth initialization.

\noindent\textbf{Depth model compatibility.} RoMeO uses DPT-Hybrid as the monocular depth prior. To verify that RoMeO is compatible with other depth models, we apply the techniques of RoMeO to the Droid-Metric3D~\citep{yin2023metric3d} baseline.
We perform the same training and evaluation as for the DPT-based RoMeO except that the monocular depth model is changed to Metric3D. Comparing row 3 of Table~\ref{tab:ablation} (\emph{DPT $\rightarrow$ Metric3D}) with the Droid-Metric3D baseline in Table~\ref{tab:result_VO_main}, we see that RoMeO is also compatible with other depth models and can effectively improve the general robustness. Meanwhile, changing DPT to Metric3D increases both ATE and RTE. This result validates our depth model choice. 

\begin{table}[!htb]
  \centering
  \caption{\textbf{Efficiency of RoMeO and a fast variant.} The speed is measured on the same RTX-3090 GPU.}
  \label{tab:efficiency}
  \resizebox{1\columnwidth}{!}{
  \begin{tabular}{@{}l|c|c|c|c@{}}
    \toprule
    \multirow{2}{*}{Model} & \multicolumn{2}{c|}{KITTI Odometry}  & \multicolumn{2}{c}{TUM-RGBD} \\
    \cmidrule{2-3} \cmidrule{4-5}
    & RTE (m) / ATE (m) & FPS & RTE (m) / ATE (m) & FPS \\
    \midrule
    No depth prior   & 47.53/137.33  & \underline{5.33} & 0.116/0.551 & \underline{10.77} \\
    RoMeO-VO & \textbf{2.71/3.81} & 3.54 & \textbf{0.067/0.091} & 7.65 \\
    RoMeO-VO-fast & \underline{4.23/6.89} & \textbf{6.28}  & \underline{0.077/0.140} & \textbf{20.57}\\
    \bottomrule
  \end{tabular}
  }
\end{table}

\noindent\textbf{Efficiency.} RoMeO introduces two depth models to improve the performance and enable metric scale VO. Here we show that the overhead of introducing depth priors is small. We also demonstrate a fast version of RoMeO which is even faster than the base VO system without depth priors, while maintaining most performance gain. 
Tab.~\ref{tab:efficiency} compares the original and the fast version of RoMeO with the baseline without depth priors. RoMeO-VO improves the accuracy significantly with marginal overhead. To create the fast version (RoMeO-VO-fast), we first reduce the input resolution, e.g., from 320*512 to 224*448 on KITTI Odometry
and from 240*320 to 192*256 at TUM-RGBD. Then, we disable the monocular depth initialization on non-keyframes. To ensure the scale consistency of the non-keyframes, we use the depth of the nearest keyframe to initialize the non-keyframes. These simple strategies preserve most of the accuracy improvement while making RoMeO-VO-fast even faster than the base VO system without depth priors, achieving twice the speed of the base system on TUM-RGBD. 




\section{Conclusion}
We propose \emph{RoMeO}, a robust visual odometry (VO) system that can return metric-scale trajectories from monocular RGB videos without 3D sensors. RoMeO utilizes pre-trained monocular and multi-view depth models as effective priors for VO. It adaptively selects accurate depth outputs to regularize BA, creating a positive feedback loop that enhances optimization robustness against local minima. Noise-augmented training is introduced to fully adapt VO networks to depth-enhanced inputs while maintaining robustness to prior noise. RoMeO consistently and significantly outperforms SOTA across 6 diverse zero-shot datasets. The performance gain also transfers to the full SLAM system.  In terms of limitations, RoMeO currently uses separate models/hyperparameters for indoor and outdoor scenes. An interesting future direction would be to develop better depth estimation methods that are lightweight, and generalize to both indoor and outdoor scenes with a single model and hyperparameter setting. 

{
    \small
    \bibliographystyle{ieeenat_fullname}
    \bibliography{main}

\begin{thebibliography}{41}
\providecommand{\natexlab}[1]{#1}
\providecommand{\url}[1]{\texttt{#1}}
\expandafter\ifx\csname urlstyle\endcsname\relax
  \providecommand{\doi}[1]{doi: #1}\else
  \providecommand{\doi}{doi: \begingroup \urlstyle{rm}\Url}\fi

\bibitem[DPT()]{DPT_hyperparam}
Dpt default scale and shift.

\bibitem[Bae et~al.(2022)Bae, Budvytis, and Cipolla]{bae2022multi}
Gwangbin Bae, Ignas Budvytis, and Roberto Cipolla.
\newblock Multi-view depth estimation by fusing single-view depth probability with multi-view geometry.
\newblock In \emph{Proceedings of the IEEE/CVF Conference on Computer Vision and Pattern Recognition}, pages 2842--2851, 2022.

\bibitem[Bhat et~al.(2023)Bhat, Birkl, Wofk, Wonka, and M{\"u}ller]{bhat2023zoedepth}
Shariq~Farooq Bhat, Reiner Birkl, Diana Wofk, Peter Wonka, and Matthias M{\"u}ller.
\newblock Zoedepth: Zero-shot transfer by combining relative and metric depth.
\newblock \emph{arXiv preprint arXiv:2302.12288}, 2023.

\bibitem[Burri et~al.(2016)Burri, Nikolic, Gohl, Schneider, Rehder, Omari, Achtelik, and Siegwart]{Burri25012016}
Michael Burri, Janosch Nikolic, Pascal Gohl, Thomas Schneider, Joern Rehder, Sammy Omari, Markus~W Achtelik, and Roland Siegwart.
\newblock The euroc micro aerial vehicle datasets.
\newblock \emph{The International Journal of Robotics Research}, 2016.

\bibitem[Campos et~al.(2021)Campos, Elvira, Rodr{\'\i}guez, Montiel, and Tard{\'o}s]{campos2021orb}
Carlos Campos, Richard Elvira, Juan J~G{\'o}mez Rodr{\'\i}guez, Jos{\'e}~MM Montiel, and Juan~D Tard{\'o}s.
\newblock Orb-slam3: An accurate open-source library for visual, visual--inertial, and multimap slam.
\newblock \emph{IEEE Transactions on Robotics}, 37\penalty0 (6):\penalty0 1874--1890, 2021.

\bibitem[Cao et~al.(2022)Cao, Ren, and Fu]{cao2022mvsformer}
Chenjie Cao, Xinlin Ren, and Yanwei Fu.
\newblock Mvsformer: Multi-view stereo by learning robust image features and temperature-based depth.
\newblock \emph{Transactions on Machine Learning Research}, 2022.

\bibitem[Cheng et~al.(2022)Cheng, Yang, Pu, and Guo]{cheng2022region}
Junda Cheng, Xin Yang, Yuechuan Pu, and Peng Guo.
\newblock Region separable stereo matching.
\newblock \emph{IEEE Transactions on Multimedia}, 25:\penalty0 4880--4893, 2022.

\bibitem[Cheng et~al.(2024{\natexlab{a}})Cheng, Xu, Guo, and Yang]{cheng2024coatrsnet}
Junda Cheng, Gangwei Xu, Peng Guo, and Xin Yang.
\newblock Coatrsnet: Fully exploiting convolution and attention for stereo matching by region separation.
\newblock \emph{International Journal of Computer Vision}, 132\penalty0 (1):\penalty0 56--73, 2024{\natexlab{a}}.

\bibitem[Cheng et~al.(2024{\natexlab{b}})Cheng, Yin, Wang, Chen, Wang, and Yang]{cheng2024adaptive}
Junda Cheng, Wei Yin, Kaixuan Wang, Xiaozhi Chen, Shijie Wang, and Xin Yang.
\newblock Adaptive fusion of single-view and multi-view depth for autonomous driving.
\newblock In \emph{Proceedings of the IEEE/CVF Conference on Computer Vision and Pattern Recognition}, pages 10138--10147, 2024{\natexlab{b}}.

\bibitem[Cheng et~al.(2025)Cheng, Liu, Xu, Wang, Zhang, Deng, Zang, Chen, Cai, and Yang]{cheng2025monster}
Junda Cheng, Longliang Liu, Gangwei Xu, Xianqi Wang, Zhaoxing Zhang, Yong Deng, Jinliang Zang, Yurui Chen, Zhipeng Cai, and Xin Yang.
\newblock Monster: Marry monodepth to stereo unleashes power.
\newblock \emph{arXiv preprint arXiv:2501.08643}, 2025.

\bibitem[Engel et~al.(2014)Engel, Sch{\"o}ps, and Cremers]{engel2014lsd}
Jakob Engel, Thomas Sch{\"o}ps, and Daniel Cremers.
\newblock Lsd-slam: Large-scale direct monocular slam.
\newblock In \emph{European conference on computer vision}, pages 834--849. Springer, 2014.

\bibitem[Engel et~al.(2017)Engel, Koltun, and Cremers]{engel2017direct}
Jakob Engel, Vladlen Koltun, and Daniel Cremers.
\newblock Direct sparse odometry.
\newblock \emph{IEEE transactions on pattern analysis and machine intelligence}, 40\penalty0 (3):\penalty0 611--625, 2017.

\bibitem[Feng et~al.(2024)Feng, Cheng, Jia, Liu, Xu, and Yang]{feng2024mc}
Miaojie Feng, Junda Cheng, Hao Jia, Longliang Liu, Gangwei Xu, and Xin Yang.
\newblock Mc-stereo: Multi-peak lookup and cascade search range for stereo matching.
\newblock In \emph{2024 International Conference on 3D Vision (3DV)}, pages 344--353. IEEE, 2024.

\bibitem[Forster et~al.(2015)Forster, Carlone, Dellaert, and Scaramuzza]{forster2015imu}
Christian Forster, Luca Carlone, Frank Dellaert, and Davide Scaramuzza.
\newblock Imu preintegration on manifold for efficient visual-inertial maximum-a-posteriori estimation.
\newblock Technical report, 2015.

\bibitem[Geiger et~al.(2012)Geiger, Lenz, and Urtasun]{Geiger2012CVPR}
Andreas Geiger, Philip Lenz, and Raquel Urtasun.
\newblock Are we ready for autonomous driving? the kitti vision benchmark suite.
\newblock In \emph{Conference on Computer Vision and Pattern Recognition (CVPR)}, 2012.

\bibitem[Handa et~al.(2014)Handa, Whelan, McDonald, and Davison]{handa2014benchmark}
Ankur Handa, Thomas Whelan, John McDonald, and Andrew~J Davison.
\newblock A benchmark for rgb-d visual odometry, 3d reconstruction and slam.
\newblock In \emph{2014 IEEE international conference on Robotics and automation (ICRA)}, pages 1524--1531. IEEE, 2014.

\bibitem[Hu et~al.(2024)Hu, Yin, Zhang, Cai, Long, Chen, Wang, Yu, Shen, and Shen]{hu2024metric3d}
Mu Hu, Wei Yin, Chi Zhang, Zhipeng Cai, Xiaoxiao Long, Hao Chen, Kaixuan Wang, Gang Yu, Chunhua Shen, and Shaojie Shen.
\newblock Metric3d v2: A versatile monocular geometric foundation model for zero-shot metric depth and surface normal estimation.
\newblock \emph{arXiv preprint arXiv:2404.15506}, 2024.

\bibitem[Kendall et~al.(2015)Kendall, Grimes, and Cipolla]{kendall2015posenet}
Alex Kendall, Matthew Grimes, and Roberto Cipolla.
\newblock Posenet: A convolutional network for real-time 6-dof camera relocalization.
\newblock In \emph{Proceedings of the IEEE international conference on computer vision}, pages 2938--2946, 2015.

\bibitem[Mur-Artal et~al.(2015)Mur-Artal, Montiel, and Tardos]{mur2015orb}
Raul Mur-Artal, Jose Maria~Martinez Montiel, and Juan~D Tardos.
\newblock Orb-slam: a versatile and accurate monocular slam system.
\newblock \emph{IEEE transactions on robotics}, 31\penalty0 (5):\penalty0 1147--1163, 2015.

\bibitem[Ranftl et~al.(2020)Ranftl, Lasinger, Hafner, Schindler, and Koltun]{ranftl2020towards}
Ren{\'e} Ranftl, Katrin Lasinger, David Hafner, Konrad Schindler, and Vladlen Koltun.
\newblock Towards robust monocular depth estimation: Mixing datasets for zero-shot cross-dataset transfer.
\newblock \emph{IEEE transactions on pattern analysis and machine intelligence}, 44\penalty0 (3):\penalty0 1623--1637, 2020.

\bibitem[Ranftl et~al.(2021)Ranftl, Bochkovskiy, and Koltun]{ranftl2021vision}
Ren{\'e} Ranftl, Alexey Bochkovskiy, and Vladlen Koltun.
\newblock Vision transformers for dense prediction.
\newblock In \emph{Proceedings of the IEEE/CVF international conference on computer vision}, pages 12179--12188, 2021.

\bibitem[Rosinol et~al.(2020)Rosinol, Abate, Chang, and Carlone]{rosinol2020kimera}
Antoni Rosinol, Marcus Abate, Yun Chang, and Luca Carlone.
\newblock Kimera: an open-source library for real-time metric-semantic localization and mapping.
\newblock In \emph{2020 IEEE International Conference on Robotics and Automation (ICRA)}, pages 1689--1696. IEEE, 2020.

\bibitem[Schops et~al.(2019)Schops, Sattler, and Pollefeys]{schops2019bad}
Thomas Schops, Torsten Sattler, and Marc Pollefeys.
\newblock Bad slam: Bundle adjusted direct rgb-d slam.
\newblock In \emph{Proceedings of the IEEE/CVF Conference on Computer Vision and Pattern Recognition}, pages 134--144, 2019.

\bibitem[Sturm et~al.(2012)Sturm, Engelhard, Endres, Burgard, and Cremers]{sturm2012benchmark}
J{\"u}rgen Sturm, Nikolas Engelhard, Felix Endres, Wolfram Burgard, and Daniel Cremers.
\newblock A benchmark for the evaluation of rgb-d slam systems.
\newblock In \emph{2012 IEEE/RSJ international conference on intelligent robots and systems}, pages 573--580. IEEE, 2012.

\bibitem[Tateno et~al.(2017)Tateno, Tombari, Laina, and Navab]{tateno2017cnn}
Keisuke Tateno, Federico Tombari, Iro Laina, and Nassir Navab.
\newblock Cnn-slam: Real-time dense monocular slam with learned depth prediction.
\newblock In \emph{Proceedings of the IEEE conference on computer vision and pattern recognition}, pages 6243--6252, 2017.

\bibitem[Teed and Deng(2018)]{teed2018deepv2d}
Zachary Teed and Jia Deng.
\newblock Deepv2d: Video to depth with differentiable structure from motion.
\newblock \emph{arXiv preprint arXiv:1812.04605}, 2018.

\bibitem[Teed and Deng(2020)]{teed2020raft}
Zachary Teed and Jia Deng.
\newblock Raft: Recurrent all-pairs field transforms for optical flow.
\newblock In \emph{Computer Vision--ECCV 2020: 16th European Conference, Glasgow, UK, August 23--28, 2020, Proceedings, Part II 16}, pages 402--419. Springer, 2020.

\bibitem[Teed and Deng(2021)]{teed2021droid}
Zachary Teed and Jia Deng.
\newblock Droid-slam: Deep visual slam for monocular, stereo, and rgb-d cameras.
\newblock \emph{Advances in neural information processing systems}, 34:\penalty0 16558--16569, 2021.

\bibitem[Teed et~al.(2024)Teed, Lipson, and Deng]{teed2024deep}
Zachary Teed, Lahav Lipson, and Jia Deng.
\newblock Deep patch visual odometry.
\newblock \emph{Advances in Neural Information Processing Systems}, 36, 2024.

\bibitem[Wang et~al.(2017)Wang, Schworer, and Cremers]{wang2017stereo}
Rui Wang, Martin Schworer, and Daniel Cremers.
\newblock Stereo dso: Large-scale direct sparse visual odometry with stereo cameras.
\newblock In \emph{Proceedings of the IEEE International Conference on Computer Vision}, pages 3903--3911, 2017.

\bibitem[Wang et~al.(2020)Wang, Zhu, Wang, Hu, Qiu, Wang, Hu, Kapoor, and Scherer]{wang2020tartanair}
Wenshan Wang, Delong Zhu, Xiangwei Wang, Yaoyu Hu, Yuheng Qiu, Chen Wang, Yafei Hu, Ashish Kapoor, and Sebastian Scherer.
\newblock Tartanair: A dataset to push the limits of visual slam.
\newblock In \emph{2020 IEEE/RSJ International Conference on Intelligent Robots and Systems (IROS)}, pages 4909--4916. IEEE, 2020.

\bibitem[Wang et~al.(2021)Wang, Hu, and Scherer]{wang2021tartanvo}
Wenshan Wang, Yaoyu Hu, and Sebastian Scherer.
\newblock Tartanvo: A generalizable learning-based vo.
\newblock In \emph{Conference on Robot Learning}, pages 1761--1772. PMLR, 2021.

\bibitem[Wenzel et~al.(2021)Wenzel, Wang, Yang, Cheng, Khan, von Stumberg, Zeller, and Cremers]{wenzel20214seasons}
Patrick Wenzel, Rui Wang, Nan Yang, Qing Cheng, Qadeer Khan, Lukas von Stumberg, Niclas Zeller, and Daniel Cremers.
\newblock 4seasons: A cross-season dataset for multi-weather slam in autonomous driving.
\newblock In \emph{Pattern Recognition: 42nd DAGM German Conference, DAGM GCPR 2020, T{\"u}bingen, Germany, September 28--October 1, 2020, Proceedings 42}, pages 404--417. Springer, 2021.

\bibitem[Whelan et~al.(2013)Whelan, Johannsson, Kaess, Leonard, and McDonald]{whelan2013robust}
Thomas Whelan, Hordur Johannsson, Michael Kaess, John~J Leonard, and John McDonald.
\newblock Robust real-time visual odometry for dense rgb-d mapping.
\newblock In \emph{2013 IEEE International Conference on Robotics and Automation}, pages 5724--5731. IEEE, 2013.

\bibitem[Yang et~al.(2024)Yang, Kang, Huang, Xu, Feng, and Zhao]{yang2024depth}
Lihe Yang, Bingyi Kang, Zilong Huang, Xiaogang Xu, Jiashi Feng, and Hengshuang Zhao.
\newblock Depth anything: Unleashing the power of large-scale unlabeled data.
\newblock \emph{arXiv preprint arXiv:2401.10891}, 2024.

\bibitem[Yang et~al.(2020)Yang, Stumberg, Wang, and Cremers]{yang2020d3vo}
Nan Yang, Lukas~von Stumberg, Rui Wang, and Daniel Cremers.
\newblock D3vo: Deep depth, deep pose and deep uncertainty for monocular visual odometry.
\newblock In \emph{Proceedings of the IEEE/CVF conference on computer vision and pattern recognition}, pages 1281--1292, 2020.

\bibitem[Yao et~al.(2018)Yao, Luo, Li, Fang, and Quan]{yao2018mvsnet}
Yao Yao, Zixin Luo, Shiwei Li, Tian Fang, and Long Quan.
\newblock Mvsnet: Depth inference for unstructured multi-view stereo.
\newblock In \emph{Proceedings of the European conference on computer vision (ECCV)}, pages 767--783, 2018.

\bibitem[Yin et~al.(2023)Yin, Zhang, Chen, Cai, Yu, Wang, Chen, and Shen]{yin2023metric3d}
Wei Yin, Chi Zhang, Hao Chen, Zhipeng Cai, Gang Yu, Kaixuan Wang, Xiaozhi Chen, and Chunhua Shen.
\newblock Metric3d: Towards zero-shot metric 3d prediction from a single image.
\newblock In \emph{Proceedings of the IEEE/CVF International Conference on Computer Vision}, pages 9043--9053, 2023.

\bibitem[Zhang et~al.(2024)Zhang, Cheng, Xu, Wang, Zhang, and Yang]{zhang2024leveraging}
Zhaoxing Zhang, Junda Cheng, Gangwei Xu, Xiaoxiang Wang, Can Zhang, and Xin Yang.
\newblock Leveraging consistent spatio-temporal correspondence for robust visual odometry.
\newblock \emph{arXiv preprint arXiv:2412.16923}, 2024.

\bibitem[Zhu et~al.(2025)Zhu, Xiang, Wang, Liu, Wang, Zhang, Guo, and Yang]{zhu2025svdc}
Xuan Zhu, Jijun Xiang, Xianqi Wang, Longliang Liu, Yu Wang, Hong Zhang, Fei Guo, and Xin Yang.
\newblock Svdc: Consistent direct time-of-flight video depth completion with frequency selective fusion.
\newblock \emph{arXiv preprint arXiv:2503.01257}, 2025.

\bibitem[Zubizarreta et~al.(2020)Zubizarreta, Aguinaga, and Montiel]{zubizarreta2020direct}
Jon Zubizarreta, Iker Aguinaga, and Jose Maria~Martinez Montiel.
\newblock Direct sparse mapping.
\newblock \emph{IEEE Transactions on Robotics}, 36\penalty0 (4):\penalty0 1363--1370, 2020.

\end{thebibliography}
}

\clearpage

\appendix

\section{Appendix}\label{appdx:detailed_results}

As shown in Tables ~\ref{tab:kitti_vo_individual} to~\ref{tab:ETH3D_vo_individual}, RoMeO outperforms the baselines on most individual sequences of each dataset. Unlike previous learning-based methods that often perform worse than conventional methods when they do not fail, RoMeO performs better and improves over SOTA on most sequences. 



  \begin{table*}[hb]
  \caption{\textbf{VO (top) and SLAM (bottom) results on the individual sequences of \emph{KITTI Odometry}}\citep{Geiger2012CVPR}.}
  \centering
  \resizebox{.98\linewidth}{!}{
  \label{tab:kitti_vo_individual}
  \begin{tabular}{@{}l|c|c|c|c|c|c|c|c|c|c@{}}
    \toprule
     VO result & 00  & 03 & 04 & 05 & 06 & 07 & 08 & 09 & 10 & average \\
    \midrule
    ORB-SLAM3 (VO)  & 49.43/182.37  & \textbf{0.64}/148.71 & 1.81/120.53 & 33.22/146.49 & 54.28/125.27 & 16.20/81.15 & 52.58/249.35 & 46.61/208.96 & 7.13/200.24 & 29.10/163.25 \\
    DSO  & 48.04/180.58 & 0.80/142.58 & \textbf{0.36}/73.46 & 48.45/143.82 & 57.59/136.59 & 53.67/66.80 & 113.02/230.02 & 92.19/218.11 & 11.03/196.32 & 47.23/154.25 \\
    \midrule
    DeepV2d  & 101.65/173.50 & 7.15/150.03 & 4.08/98.86 & 27.05/142.89 & 7.39/120.58 & 8.70/80.53 & 18.91/236.56 & 10.13/199.01 & 14.77/188.74 & 22.20/154.52 \\
    TrianFlow & 96.43/186.91 & 10.26/163.78 & 6.05/109.07 & 52.69/154.90 & 39.00/132.37 & 15.69/88.18 & 56.63/256.00 & 72.59/218.89 & 29.35/203.44 & 42.07/168.17 \\
    TartanVO  & 63.84/63.91 & 7.71/21.56 & 2.89/49.45 & 54.61/56.57 & 24.67/47.10 & 19.29/19.95 & 59.55/59.56 & 32.61/59.23 & 25.04/30.41 & 32.25/45.30 \\
    DROID-VO & 109.00/111.17 & 5.57/150.5 & 1.05/108.5 & 60.37/121.25 & 38.03/123.50 & 21.41/74.33 & 105.64/159.03 & 73.04/193.30 & 13.77/194.52 & 47.53/137.33\\   
    DROID-Metric3d & \textbf{4.38/5.35} & 1.01/5.90 & 0.74/2.11 & 5.27/\textbf{7.72} & 3.19/3.76 & 8.84/11.27 & 6.59/8.09 & 2.99/3.00 & 2.55/2.97 & 3.95/5.57 \\
    DPVO & 110.93/111.78 & 1.67/156.68 & 1.05/107.81 & 55.89/127.76 & 59.84/118.55 & 18.21/69.27 & 97.19/189.61 & 60.25/196.83 & 9.37/184.23 & 46.04/140.28 \\
    RoMeO-VO (ours) & 4.78/5.88 & 0.98/\textbf{2.22} & 0.62/\textbf{1.77} & \textbf{3.35}/7.84 & \textbf{1.93/1.97} & \textbf{3.95/3.96} & \textbf{4.46/6.31} & \textbf{2.65/2.73} & \textbf{1.63/1.67} & \textbf{2.71/3.81}\\
  \bottomrule
  \end{tabular}
  }
 
 \vspace{+1em}
 
  \centering
  \resizebox{.98\linewidth}{!}{
  \label{tab:4seasons_vo_individual}
  \begin{tabular}{@{}l|c|c|c|c|c|c|c|c|c|c@{}}
    \toprule
     SLAM result & 00  & 03 & 04 & 05 & 06 & 07 & 08 & 09 & 10 & average \\
    \midrule
    ORB-SLAM3   & 9.27/189.54  & \textbf{0.58}/144.83 & 1.69/119.27 & 5.73/162.10 & 16.21/136.77 & 2.71/83.21 & 51.01/246.46 & 53.70/210.03 & 6.88/195.34 & 16.42/165.28 \\
    Droid-SLAM  & 74.41/167.26  & 7.37/156.09 & \textbf{0.43}/105.63 & 60.14/140.15 & 38.45/126.85 & 20.66/81.38 & 69.33/228.68 & 65.81/207.15 & 15.48/189.84 & 39.12/155.89 \\
    RoMeO-SLAM (ours) & \textbf{4.74/5.75} & 0.95/\textbf{2.18} & 0.57/\textbf{1.76} & \textbf{3.28/7.28} & \textbf{1.88/1.90} & 3.55/\textbf{3.78} & \textbf{4.52/6.45} & \textbf{2.63/2.69} & \textbf{1.65/1.71} & \textbf{2.64/3.72}\\
  \bottomrule
  \end{tabular}
  }
\end{table*}

\begin{table*}[hb]
  \caption{
  \textbf{VO (top) and SLAM (bottom) results on the individual sequences of \emph{4Seasons}}~\citep{wenzel20214seasons}.
  }
  \centering
  \resizebox{.98\linewidth}{!}{
  \label{tab:4Seasons_vo_individual}

  \begin{tabular}{@{}l|c|c|c|c|c|c|c|c@{}}
    \toprule
     VO result & business\underline{~}campus & old\underline{~}town & parking\underline{~}garage & neighborhood & office\underline{~}loop & city\underline{~}loop & countryside & average \\
    \midrule
    ORB-SLAM3 (VO)   & 88.21/163.57  & 15.10/201.25 & 4.58/18.54 & 74.95/160.62 & 48.32/80.13 & -- & -- & -- \\
    DSO  & 143.23/163.79 & -- & -- & -- & 72.52/166.32 & 56.31/69.72 & 56.72/102.61 & -- \\
    \midrule
    DeepV2d  & 62.03/1547.58 & 36.01/204.19 & 18.81/19.06 & 98.33/129.06 & 88.50/173.16 & 105.75/137.01 & 113.49/203.21 & 74.70/344.75 \\
    TrianFlow & 20.92/161.39 & 37.05/216.35 & 13.29/19.02 & 27.80/143.48 & 31.94/172.04 & 63.32/116.43 & 88.75/224.53 & 40.44/150.46 \\
    TartanVO  & 80.37/133.28 & 69.18/69.93 & 10.69/37.92 & 88.83/136.69 & 68.75/68.81 & 17.69/\textbf{31.23} & 79.23/94.03 & 59.25/81.70 \\
    DROID-VO & 23.06/161.54 & 8.85/209.24 & 14.32/19.15 & 12.19/143.04 & 60.71/171.98 & 115.03/116.57 & 177.93/225.24 & 58.87/149.54 \\   
    DROID-Metric3d & 94.33/104.09 & 212.85/217.34 & 12.92/13.78 & 98.37/135.31 & 142.19/170.17 & 96.77/116.40 & 220.63/225.13 & 125.44/140.32 \\
    DPVO & 8.19/161.27 & 19.68/161.88 & \textbf{0.98/14.89} & \textbf{7.68}/140.77 & \textbf{21.54}/170.91 & \textbf{1.96}/115.41 & \textbf{9.63}/224.38 & \textbf{9.95}/141.36 \\
    RoMeO-VO (ours) & \textbf{7.41/13.74} & \textbf{3.97/27.87} & 10.86/35.60 & 11.89/\textbf{18.13} & 38.78/\textbf{72.22} & 24.59/77.69 & 39.62/\textbf{52.67} & 19.59/\textbf{42.56}\\
  \bottomrule
  \end{tabular}
  }

\vspace{+1em}

\centering
  \resizebox{.98\linewidth}{!}{

  \begin{tabular}{@{}l|c|c|c|c|c|c|c|c@{}}
    \toprule
     SLAM result & business\underline{~}campus & old\underline{~}town & parking\underline{~}garage & neighborhood & office\underline{~}loop & city\underline{~}loop & countryside & average \\
    \midrule
    ORB-SLAM3  & 100.35/163.21  & 14.02/206.53 & \textbf{2.79/16.83} & 78.82/163.25 & 48.16/82.07 & -- & -- & -- \\
    Droid-SLAM  & 23.55/152.99  & 8.54/205.85 & 14.38/16.40 & 14.28/134.05 & 59.04/158.63 & 113.54/116.52 & 173.50/224.93 & 58.12/144.20 \\
    RoMeO-SLAM (ours) & \textbf{8.32/13.84} & \textbf{4.04/27.49} & 10.51/36.03 & \textbf{11.59/20.49} & \textbf{42.05/77.89} & \textbf{26.87/80.26} & \textbf{39.37/53.45} & \textbf{20.39/44.20}\\
  \bottomrule
  \end{tabular}
  }
\end{table*}



\begin{table*}[h]
  \caption{\textbf{VO (top) and SLAM (bottom) results on the individual sequences of \emph{Cambridge}}~\citep{kendall2015posenet}. 
  }
  \centering
  \resizebox{.98\linewidth}{!}{
  \label{tab:Cambridge_vo_individual}
  \begin{tabular}{@{}l|c|c|c|c|c|c|c@{}}
    \toprule
     VO result & GreatCourt  & KingsCollege & OldHospital & ShopFacade & StMarysChurch & Street & average \\
    \midrule
    ORB-SLAM3 (VO)  & --  & -- & -- & -- & -- & -- & -- \\
    DSO  & -- & -- & -- & --  & -- & -- & -- \\
    \midrule
    DeepV2d  & 27.55/33.38 &15.27/29.24 & 8.17/10.78 & 7.18/8.73 & 12.04/15.75 & 93.06/109.12 & 27.21/34.50 \\
    TrianFlow & 27.93/32.93 & 21.42/30.15 & 7.30/10.84 & 5.97/8.96 & 11.93/16.05 & 89.52/108.67 & 27.35/34.60\\
    TartanVO & 28.09/34.31 & 12.81/16.92 & 6.60/7.54 & 4.01/4.22 & 10.60/11.82 & 54.28/\textbf{70.60} & 19.40/24.23  \\
    DROID-VO & 14.79/32.48 & 1.11/29.2 & \textbf{0.94}/11.92 & 0.53/8.09 & 4.35/14.99 & 59.66/105.34 & 13.56/33.67  \\   
    DROID-Metric3d & 13.87/40.80 & 0.81/18.08 & 1.31/\textbf{7.53} & 0.75/12.98 & 4.41/21.51 & 60.16/185.23 & 13.55/47.68 \\
    DPVO & 37.28/38.34 & \textbf{0.18/16.12} & 4.07/10.86 & \textbf{0.12}/8.68 & 5.27/18.09 & 48.39/124.06 & 15.89/36.03 \\
    RoMeO-VO (ours) & \textbf{12.78/26.79} & 0.40/19.2 & 1.14//8.56 & 0.52/\textbf{3.27} & \textbf{4.28/10.28} & \textbf{40.68}/72.76 & \textbf{9.96/23.47}
\\
  \bottomrule
  \end{tabular}
  }

\vspace{+1em}

  \centering
  \resizebox{.98\linewidth}{!}{
  \begin{tabular}{@{}l|c|c|c|c|c|c|c@{}}
    \toprule
     SLAM result & GreatCourt  & KingsCollege & OldHospital & ShopFacade & StMarysChurch & Street & average \\
    \midrule
    ORB-SLAM3  & --  & -- & -- & -- & -- & -- & -- \\
    Droid-SLAM & 13.78/32.49 & 0.34/29.23 & \textbf{0.89}/10.58 & 0.51/8.25  & \textbf{4.17}/15.14 & 54.50/103.08 & 12.37/33.13 \\
    RoMeO-SLAM (ours) & \textbf{12.70/26.73} & \textbf{0.30/18.70} & 1.06/\textbf{8.44} & \textbf{0.49/3.23} & 4.25/\textbf{10.35} & \textbf{40.61/69.66} & \textbf{9.90/23.23}
\\
  \bottomrule
  \end{tabular}
  }
  
\end{table*}

\begin{table*}[tb]
  \caption{\textbf{VO (top) and SLAM (bottom) results on the individual sequences of \emph{EuRoC}~\citep{Burri25012016}}. 
  }
  \centering
  \resizebox{.98\linewidth}{!}{
  \label{tab:Cambridge_vo_individual}
  \begin{tabular}{@{}l|c|c|c|c|c|c|c@{}}
    \toprule
     VO result & V101  & V102 & V103 & V201 & V202 & V203 & average \\
    \midrule
    ORB-SLAM3 (VO)  & \textbf{0.036}/1.045  & 0.139/4.210 & 0.713/1.108 & 1.352/1.866 & \textbf{0.047}/0.302 & 0.642/1.529 & 0.488/1.677 \\
    DSO  & 0.089/0.937 & \textbf{0.107}/3.859 & 0.903/1.236 & 0.044/1.244  & 0.132/0.361 & 1.152/1.555 & 0.404/1.532 \\
    \midrule
    DeepV2d  & 0.717/1.365 & 0.695/6.210 & 1.483/5.544 & 0.839/3.109 & 1.052/2.331 & 0.591/2.153 & 1.173/3.452 \\
    TrianFlow & 0.895/1.243 & 3.956/4.038 & 0.974/1.076 & 1.849/1.927 & 0.483/0.547 & 2.229/2.257 & 1.731/1.848\\
    TartanVO & 0.447/2.357 & 0.389/6.901 & 0.622/5.921 & 0.433/3.644 & 0.749/3.015 & 1.152/2.973 & 0.632/4.135\\
    DROID-VO & 0.103/1.203 & 0.165/1.847 & 0.158/1.061 & 0.102/1.360 & 0.115/0.992 & 0.204/1.378 & 0.141/1.307  \\   
    DROID-Metric3d & 0.054/1.234 & 0.141/\textbf{1.815} & 0.112/1.135 & 0.075/1.253 & 0.116/1.057 & 0.156/1.366 & 0.109/1.310 \\
    DPVO & 0.048/1.190 & 0.148/3.658 & \textbf{0.093/0.759} & 0.059/1.626
 & 0.051/2.361 & 0.207/1.598 & 0.101/1.865 \\
    RoMeO-VO (ours) & 0.073/\textbf{0.892} & 0.136/2.634 & 0.101/0.991 & \textbf{0.056/0.970} & 0.107/\textbf{0.132} & \textbf{0.117/1.139} & \textbf{0.098/1.126}
\\
  \bottomrule
  \end{tabular}
  }

\vspace{+1em}

  \centering
  \resizebox{.98\linewidth}{!}{
  \begin{tabular}{@{}l|c|c|c|c|c|c|c@{}}
    \toprule
     SLAM result & GreatCourt  & KingsCollege & OldHospital & ShopFacade & StMarysChurch & Street & average \\
    \midrule
    ORB-SLAM3  & 0.033/0.992 & 0.042/3.854 & 0.395/0.952 & 0.683/1.705 & 0.028/0.282 & 0.103/0.837 & 0.214/1.437 \\
    Droid-SLAM & 0.037/0.920 & \textbf{0.013/1.430} & 0.019/1.123 & 0.017/1.437  & 0.014/1.100 & \textbf{0.013}/1.418 & 0.019/1.238 \\
    RoMeO-SLAM (ours) & \textbf{0.031/0.88} & \textbf{0.013}/2.556 & \textbf{0.018/0.990} & \textbf{0.016/0.963} & \textbf{0.013/0.111} & \textbf{0.013/1.118} & \textbf{0.016/1.100}
\\
  \bottomrule
  \end{tabular}
  }
  
\end{table*}


\begin{table*}[t]
  \caption{\textbf{VO (top) and SLAM (bottom) results on the individual sequences of \emph{ETH3D SLAM}}~\citep{schops2019bad}.
  }
  \centering
  \resizebox{.98\linewidth}{!}{
  \label{tab:ETH3D_vo_individual}
  \begin{tabular}{@{}l|c|c|c|c|c|c|c|c@{}}
    \toprule
     VO result & cables  & camera\underline{~}shake & desk\underline{~}changing & einstein & planar & mannequin\underline{~}face & sfm\underline{~}lab\underline{~}room & average \\
    \midrule
    ORB-SLAM3 (VO)  & --  & -- & -- & -- & 0.021/\textbf{0.088} & 0.535/0.539 & -- & -- \\
    DSO  & -- & -- & 1.312/1.530 & -- & 0.686/0.689 & 0.490/0.491 & -- & -- \\
    \midrule
    DeepV2d & 0.172/0.130 & 0.155/1.394 & 0.943/1.065 & 0.407/0.467 & 0.424/0.430 & 0.119/\textbf{0.124} & 0.051/0.466 & 0.324/0.582 \\
    TrianFlow  & 0.262/0.601 & 0.152/0.206 & 1.373/1.485 & 1.037/1.044 & 0.729/0.730 & 0.447/0.475 & 0.941/1.535 & 0.706/0.868 \\
    TartanVO & 0.132/24.768 & 0.152/9.070 & 0.851/51.524 & 0.831/33.775 & 0.506/27.802 & 0.197/17.822 & 0.277/19.152 & 0.421/26.273 \\
    DROID-VO & \textbf{0.013}/0.261 & 0.140/0.194 & 0.186/1.090 & 0.824/0.876 & 0.010/0.126 & 0.329/0.366 & 1.066/1.485 & 0.367/0.628\\   
    DROID-Metric3d & 0.262/0.364 & 0.137/0.525 & 1.139/2.668 & 0.612/1.023 & 0.354/0.492 & 0.290/2.139 & 0.146/0.328 & 0.420/1.077 \\
    DPVO & 0.020/0.551 & 0.142/0.332 & 1.192/2.383 & 0.022/0.358 & 0.022/0.289 & 0.003/0.134 & \textbf{0.021}/0.479 & 0.203/0.646\\
    RoMeO-VO (ours) & 0.027/\textbf{0.042} & \textbf{0.048/0.131} & \textbf{0.028/0.396} & \textbf{0.008/0.340} & \textbf{0.006}/0.550 & \textbf{0.002}/0.158 & 0.036/\textbf{0.050} & \textbf{0.022/0.238}\\
  \bottomrule
  \end{tabular}
  }
  \vspace{+1em}

  \centering
  \resizebox{.98\linewidth}{!}{
  \begin{tabular}{@{}l|c|c|c|c|c|c|c|c@{}}
    \toprule
     SLAM result & cables  & camera\underline{~}shake & desk\underline{~}changing & einstein & planar & mannequin\underline{~}face & sfm\underline{~}lab\underline{~}room & average \\
    \midrule
    ORB-SLAM3   & --  & -- & -- & -- & 0.012/0.224 & 0.287/0.335 & -- & -- \\
    Droid-SLAM  & \textbf{0.005/0.007} & 0.049/0.049 & \textbf{0.004/0.005} & 0.004/\textbf{0.005} & \textbf{0.002/0.002} & \textbf{0.002/0.004} & \textbf{0.005/0.008} & 0.010/\textbf{0.011} \\
    RoMeO-SLAM (ours) & 0.017/0.028 & \textbf{0.015/0.032} & \textbf{0.004}/0.118 & \textbf{0.002}/0.108 & \textbf{0.002}/0.004 & \textbf{0.002}/0.060 & 0.016/0.070 & \textbf{0.008}/0.060\\
  \bottomrule
  \end{tabular}
  }
\end{table*}

\begin{table*}[tb]
  \caption{\textbf{VO (top) and SLAM (bottom) results on the individual sequences of \emph{TUM-RGBD}}~\citep{sturm2012benchmark}.
  }
  \centering
  \resizebox{.98\linewidth}{!}{
  \label{tab:TUM_vo_individual}

  \begin{tabular}{@{}l|c|c|c|c|c|c|c|c|c|c@{}}
    \toprule
     VO result & 360  & desk & desk2 & floor & plant & room & rpy & teddy & xyz & average \\
    \midrule
    ORB-SLAM3 (VO)  & --  & \textbf{0.017}/0.065 & -- & -- & \textbf{0.034}/0.492 & -- & -- & -- & 0.009/0.061 & -- \\
    DSO  & -- & 0.405/0.535 & 0.322/0.818 & \textbf{0.041}/0.119 & 0.108/0.302 & 0.800/0.850 & -- & -- & 0.058/0.073 & -- \\
    \midrule
    DeepV2d  & 0.144/0.250 & 0.105/0.327 & 0.321/0.578 & 0.628/1.514
 & 0.217/0.330 & 0.215/0.250 & 0.046/0.115 & 0.294/0.312 & 0.051/0.055 & 0.225/0.415 \\
    TrianFlow  & 0.187/0.187 & 0.526/0.698 & 0.483/0.764
 & 0.739/0.741 & 0.388/0.610 & 0.884/0.929 & 0.050/0.158 & 0.554/0.762 & 0.182/0.236 & 0.444/0.565 \\
    TartanVO  & 0.160/7.212 & 0.478/23.082 & 0.539/25.080 & 0.348/20.811 & 0.395/10.374 & 0.417/19.490 & 0.050/10.289 & 0.329/19.791 & 0.160/14.095 & 0.320/16.692 \\
    DROID-VO & 0.141/0.556 & 0.064/0.335 & 0.078/1.001 & 0.063/0.152 & 0.041/0.631 & 0.393/0.704 & 0.030/0.811 & 0.221/0.714 & 0.017/0.057 & 0.116/0.551\\   
    DROID-Metric3d & 0.222/0.263 & 0.139/0.260 & 0.075/0.109 & 0.195/0.769 & 0.271/0.470 & 0.264/0.378 & 0.019/\textbf{0.019} & 0.507/0.669 & 0.019/0.036 & 0.190/0.330 \\
    DPVO & 0.156/ 0.174 & 0.034/0.258 & 0.050/0.216 & 0.183/0.270 & 0.034/0.575 & 0.383/0.400 & 0.038/0.104 & 0.073/0.852 & 0.012/0.069 &  0.107/0.324\\
    RoMeO-VO (ours) & \textbf{0.084/0.085} & 0.049/\textbf{0.051} & \textbf{0.038/0.059} & 0.103/\textbf{0.104} & 0.079/\textbf{0.133} & \textbf{0.167/0.169} & \textbf{0.015}/0.030 & \textbf{0.069/0.217} & \textbf{0.005/0.006} & \textbf{0.067/0.091}\\
  \bottomrule
  \end{tabular}
  }
  \vspace{+1em}

  \centering
  \resizebox{.98\linewidth}{!}{

  \begin{tabular}{@{}l|c|c|c|c|c|c|c|c|c|c@{}}
    \toprule
    SLAM result & 360  & desk & desk2 & floor & plant & room & rpy & teddy & xyz & average \\
    \midrule
    ORB-SLAM3  & --  & 0.016/0.069 & -- & -- & 0.038/0.510 & -- & -- & 0.145/0.728 & \textbf{0.005}/0.087 & -- \\
    Droid-SLAM  & 0.063/0.081  & 0.017/0.100 & \textbf{0.026}/0.152 & \textbf{0.022/0.055} & 0.016/0.291 & 0.043/0.280 & 0.023/0.023 & 0.035/0.515 & 0.010/0.022 & 0.028/0.168 \\
    RoMeO-SLAM (ours) & \textbf{0.038/0.038} & \textbf{0.012/0.083} & 0.027/\textbf{0.080} & 0.033/0.105 & \textbf{0.011/0.014} & \textbf{0.035/0.083} & \textbf{0.012/0.019} & \textbf{0.020/0.213} & \textbf{0.005/0.005} & \textbf{0.021/0.071}\\
  \bottomrule
  \end{tabular}
  }
\end{table*}

\end{document}